\documentclass{article}
\usepackage{arxiv}
\usepackage{tikz}
\usepackage{amsmath}
\usepackage{amsfonts}
\usepackage{amssymb}
\usepackage{amsbsy}
\usepackage{amsthm}
\usepackage{caption}
\usepackage{subcaption}
\usepackage{color}
\usepackage{multirow}
\usepackage{mdframed}
\usepackage{adjustbox}

\newcommand{\nY}{\mathbf Y_n}

\newcommand{\iid}{i.i.d.}
\newcommand{\bbP}{\mathbb P}
\newcommand{\bbR}{\mathbb R}

\newcommand{\itn}{1\leq i \leq n}
\newcommand{\bbE}{\mathbb{E}}

\newcommand{\bfz}{\mathbf{Z}}

\newcommand{\pb}{\frac{\partial}{\partial \beta^j}}

\usepackage[pdftex,colorlinks=true,urlcolor=blue,citecolor=black,anchorcolor=black,linkcolor=black]{hyperref}

\newtheoremstyle{wsc}
{3pt}
{3pt}
{}
{}
{\bf}
{}
{.5em}
{}

\theoremstyle{wsc}
\newtheorem{theorem}{Theorem}

\newtheorem{definition}{Definition}

\begin{document}

%
%

\title{ESTIMATING STOCHASTIC POISSON INTENSITIES USING DEEP LATENT MODELS}

\author{
 Ruixin Wang \\
School of Industrial Engineering\\
Purdue University\\
West Lafayette, IN 47906, USA\\
\texttt{wang2252@purdue.edu.}
   \And
Prateek Jaiswal \\
School of Industrial Engineering\\
Purdue University\\
West Lafayette, IN 47906, USA\\
\texttt{jaiswalp@purdue.edu.}
  \And
Harsha Honnappa\\
School of Industrial Engineering\\
Purdue University\\
West Lafayette, IN 47906, USA\\
  \texttt{honnappa@purdue.edu.} \\
}

\maketitle

\section*{ABSTRACT}
We present a new method for estimating the stochastic intensity of a doubly stochastic Poisson process. Statistical and theoretical analyses of traffic traces show that these processes are appropriate models of high intensity traffic arriving at an array of service systems. The statistical estimation of the underlying latent stochastic intensity process driving the traffic model involves a rather complicated nonlinear filtering problem. We develop a novel simulation method, using deep neural networks to approximate the path measures induced by the stochastic intensity process, for solving this nonlinear filtering problem. Our simulation studies demonstrate that the method is quite accurate on both in-sample estimation and on an out-of-sample performance prediction task for an infinite server queue.

\section{INTRODUCTION}\label{sec:intro}
This paper introduces a simulation-based method for estimating the stochastic intensity process of a doubly stochastic Poisson process (DSPP), using sample path observations of the DSPP over a fixed time horizon and under the assumption of a stochastic differential equation (SDE) model of the intensity. 
 DSPPs are widely acknowledged as an appropriate model of traffic arriving at a variety of service systems, including hospitals and call centers. Specifically, multiple statistical analyses~\cite{jongbloed2001managing,avramidis2004modeling,avramidis2005modeling,maman2007uncertainty,kim2014call} show that the (estimated) index of dispersion (i.e., the ratio of the variance to the mean) of the arrival counts typically exceeds 1 at reasonable operational time-scales; for Poisson processes the index equals 1. Furthermore, the arrival intensity appears time-varying and there are temporal correlations between traffic counts across non-overlapping time intervals. These conditions strongly indicate that the traffic process is not a Poisson process with deterministic intensity. However, at smaller time-scales (on the order of inter-arrival times) it is not possible to reject the null hypothesis that the arrival counts over a fixed time interval are Poisson distributed~\cite{kim2014call}. DSPPs can model the overdispersion, temporal correlations and time-varying nature of the intensity while remaining reasonably tractable to use for performance prediction and control/optimization tasks. A rigorous definition of DSPPs is provided in the next section.
 

The expansive definition of DSPPs allows for many models of the stochastic intensity process. A simple model advocated for modeling call center traffic in~\cite{whitt1999dynamic} assumes that the uncertainty in the arrival rates is determined by a single random variable that determines the daily `busyness' level. However, as noted in~\cite{zhang2014scaling}, the static nature of the intensity model implies it cannot account for the temporal correlation structure observed in many traffic traces.~\cite{zhang2014scaling}, in turn, suggest the use of a `dynamic' (sic) intensity model. In the context of high intensity call center traffic they show, through a combination of theoretical and empirical analysis, that a Cox-Ingersoll-Ross (CIR) diffusion is appropriate. Recall that the CIR process is defined as the solution to the SDE
\begin{align}~\label{eq:cir}
	dZ(t) = (\beta - Z(t)) dt + \eta \beta^\alpha \sqrt{Z(t)} dW(t),~\forall t \geq 0
\end{align}
where $(W(t) : t \geq 0)$ is a standard Brownian motion process, $(\alpha,\beta,\eta)$ are positive constants that constitute the parameters of the model. Specifically,~\cite{zhang2014scaling} present empirical evidence that the empirical distribution of the standardized arrival counts roughly follows a standard normal distribution, in time intervals where the mean arrival counts are `large' . This empirical observation is supported by a rigorous central limit theorem (CLT) that holds for all $\alpha \in \left(0,\frac{1}{2}\right)$. Following~\cite{zhang2014scaling}, we assume that the stochastic intensity process is well-modeled by an SDE (though not necessarily~\eqref{eq:cir}). 

In practice, the stochastic intensity is {\it latent} (i.e., unobserved) and must be estimated from traffic traces. As noted in~\cite{cheng2017history}, this estimation problem is challenging. In fact, it entails the solution of a nonlinear filtering problem where the underlying stochastic intensity process can be viewed as the `signal,' and the arrival process is a noisy `observation' of the intensity. The solution of the nonlinear filtering problem depends crucially on the computation of the {\it pathwise Kallianpur-Striebel formula} (see~\cite[Ch.\,1]{van2007filtering}), which is remarkably complicated. More crucially, the computation of the filter assumes complete knowledge of the latent intensity model. In our setting, while a {\it structure} of the model might be assumed, model parameters are unknown and must be estimated themselves.
 
We present a computational method that simultaneously estimates the intensity model and solves the nonlinear filtering problem. We model the unknown drift and diffusion functions of this SDE using deep neural networks (DNNs), which are trained by maximizing a tight lower bound on the marginal log-likelihood of the traffic process. This is an instance of a so-called {\it deep latent model} (DLM); examples of such models includes variational autoencoders (VAEs) and generative adversarial networks (GANs) used to synthesize video and image samples (so-called `deep fakes') in the artificial intelligence (AI) literature~\cite[Ch.\,20]{goodfellow}. To the best of our knowledge, this method has not been developed in the context of continuously observed stochastic processes where DNN training can be rather complicated. Recent work in~\cite{tzen2019:neuralSDE,tzen2019theoretical} considers a more restrictive class of problems where the latent signal process is a diffusion over the interval $[0,1]$ and the objective is to estimate the terminal marginal distribution using observations of a random variable dependent on the terminal marginal (latent) random variable. 

In the subsequent sections we first present an overview of DSPPs in Section~2, followed by an extensive description of the statistical estimation problem and variational autoencoders in Section~3 and~4. We present our method in Section~5, where we derive the lower bound referenced above and the DNN training procedure we have developed, based on the theory of stochastic flows by~\cite{kunita1984stochastic}. Finally, in Section~6 we present simulation results that demonstrate the efficacy of our method. Specifically, we present results on a) in-sample estimation of the stochastic intensity process itself, and on b) out-of-sample `run-through' experiments for predicting performance metrics in an infinite server queue. Section~7 concludes with a summary and some commentary on future work.

\section{DOUBLY STOCHASTIC POISSON PROCESSES}
  Let $(X(t) : t \geq 0)$ be a non-decreasing $\mathbb Z_+$-valued point process, $(X|Z)$ represent the process conditioned on the stochastic process $(Z(t) : t \geq 0)$, and Poi$(\Lambda)$ represent a Poisson process with integrated intensity function $(\Lambda(t) : t \geq 0)$. Formally, a DSPP is defined as:

\begin{definition}
	Let $(Z(t) : t \geq 0)$ be a non-negative stochastic process such that with probability one $t \mapsto Z(t)$ is locally integrable. Then, $(X(t) : t \geq 0)$ is a DSPP driven by $(Z(t) : t \geq 0)$ if $(X|Z) \sim \text{Poi}(\bfz)$, where $\bfz$ is the integrated process defined as $\bfz(s,t) := \int_s^t Z(r) dr$ for any $s < t$.
\end{definition}
\noindent That is, for any set of points $\{t_0,t_1,\ldots, t_d\} \subset (0,\infty)$, where $0 < t_0 \leq t_1 \leq \cdots \leq t_d < \infty$, the finite dimensional distributions of $(X|Z)$ satisfy
\begin{align}
	\mathbb P(X(t_0) = k_0, X(t_1) = k_1, &\ldots, X(t_d) = k_d | Z_{0:t_d}\})\\ \nonumber &= \frac{\exp(-\bfz(0,t_{0}))(\bfz(0,t_{0})^{k_0}}{k_0!}\prod_{i=0}^{d-1} \frac{\exp(-\bfz(t_i,t_{i+1}))(\bfz(t_i,t_{i+1}))^{k_{i+1}-k_{i}}}{(k_{i+1}-k_i)!},
\end{align} 
where $Z_{0:t} \equiv (Z(s) : 0 \leq s \leq t)$.
Formally, the path measure induced by $(X(t) : t \geq 0)$ is defined as $\int Poi(\bfz) dP(Z)$, where $P(\cdot)$ is the path measure induced by the stochastic process $(Z(t) : t \geq 0))$, so that the finite dimensional distribution of $X(t)$ (at any fixed $t \geq 0$) satisfies
\begin{align}
	\mathbb P\left( X(t_0) = k_0,\ldots,X(t_d) = k_d \right) = \int \mathbb P(X(t_0) = k_0, X(t_1) = k_1, &\ldots, X(t_d) = k_d | Z_{0:t_{d}}\}) dP(Z_{0:t_{d}}).
\end{align}
Note that we are deliberately being less than rigorous in our description of this path measure so as to avoid a heavier notational burden that distracts from the primary message of this paper.

\section{THE STATISTICAL ESTIMATION PROBLEM}
In the setting of a stochastic differential equation (SDE) model of the intensity, the estimation problem amounts to estimating the drift and diffusion coefficients. Suppose the drift and diffusion coefficients are parameterized by $\theta$. 
To understand the complexity of the problem, consider the following formal argument for deriving the maximum (log-)likelihood estimator (MLE) of the marginal distribution of $X(t)$,
\begin{align}~\label{eq:lln}
	\log \mathbb P\left( X(t) = k \right) &= \log \int \mathbb P(X(t) = k | Z_{0:t}) dP_\theta(Z_{0:t}),
\end{align}
where $P_\theta$ is the path measure corresponding to the parameters $\theta$. Observe that computing the MLE requires differentiating with respect to $\theta$ under this path measure. There are potentially two ways of doing this. First, suppose we are able to compute the distribution of $\bfz(0,t)$ as a function of the parameters $\theta$. Then, the gradient of the log-likelihood can be computed using the score function. However, while the distribution of $\bfz$ could be computed with some effort for some instances (such as the CIR model~\eqref{eq:cir}), this is unlikely to be true for arbitrary stochastic processes. On the other hand, suppose the path measure $P_\theta$ has a Radon-Nikodym density with respect to a reference path measure $\pi_0$; that is, there exists a real-valued potential function $\Phi(Z_{0:T};\theta)$ such that $dP_\theta/d\pi_0(Z_{0:T}) \propto \exp\left(\Phi(Z_{0:T};\theta)\right)$, then the gradient can be computed by differentiating the potential function. In general, however, we are confronted by the question of the choice of an appropriate reference measure. Note that measures in infinite dimensional spaces have a strong tendency towards either singularity or equivalence, complicating this choice; for instance, standard Brownian motion is not a feasible reference measure for the CIR process. While this issue can be resolved in specific cases, we would like a method that works for arbitrary choices of the stochastic intensity process. 

While the reference measure might not be known, we can introduce another measure into~\eqref{eq:lln} that also has a density with respect to the reference measure, making it equivalent to $P_\theta$. Observe that the conditional measure $P(Z_{0:t}|X(t)=k)$ is the ``optimal" choice in the sense that we have
\begin{align}
	\label{eq:multiply}
	 \log \bbP(X(t) = k) &=\log \int \bbP(X(t) = k | Z_{0:t}) \frac{dP_\theta(Z_{0:t})}{dP(Z_{0:t}|X(t)=k)} dP(Z_{0:t}|X(t) = k)\\
	 &=\int \log\left( \bbP(X(t) = k | Z_{0:t}) \frac{dP_\theta(Z_{0:t})}{dP(Z_{0:t}|X(t)=k)} \right)dP(Z_{0:t}|X(t) = k),
\end{align}
where the second equality follows from the fact that the term inside the $\log$ is precisely $\bbP(X(t) = k)$ (and therefore a constant with respect to the conditional measure). This formal calculation shows that computing the MLE of the count process amounts to solving a complex nonlinear filtering problem to compute the conditional measure, where the unobserved stochastic intensity function should be viewed as a `signal' and the (conditionally) Poisson counts are noisy `observations' of the signal. More precisely, the Doob-Meyer decomposition of the DSPP $(X(t):t\geq 0)$ implies $X(t) = \int_0^t Z(s) ds + \eta(t)$, where $(Z(t) : t\geq 0)$ is the stochastic intensity process and $(\eta(t) : t \geq 0)$ is a martingale (see~\cite{segall1975modeling} as well). However, solving this filtering problem is remarkably hard. Observe that the density $dP(\cdot|X(t)=k)/dP_\theta(\cdot)$ is the pathwise Kallianpur-Striebel formula~\cite[Ch.\,1]{van2007filtering}.~
Solving this nonlinear filtering problem, however, is no easier than the `direct differentiation' methods for computing the MLE noted in the previous paragraph.

Revisiting the computation in~\eqref{eq:multiply}, suppose we now introduce an arbitrary (but equivalent) measure $P_{\phi,k}$ (parameterized by $\phi$ and $k$). Then, Jensen's inequality implies that
\begin{align}~\label{eq:vae}
	\log \bbP(X(t) = k) &\geq \int \log \left( \bbP(X(t) = k | Z_{0:t}) \frac{dP_\theta(Z_{0:t})}{dP_{\phi,k}(Z_{0:t})} \right) dP_{\phi,k}(Z_{0:t}).
\end{align}
While this is a lower bound, observe that the inequality can be tightened by maximizing it over both $\theta$ and $\phi$. The objective, however, is highly non-concave in these parameters and consequently we can only guarantee the computation of a local optimum. Furthermore, the choice of parameterization will, in general, imply that the class of measures being optimized over may not include the `true' measures, resulting in an approximation to the filtering distribution. Therefore, this procedure of optimizing over path measures is an example of {\it approximate inference}, used extensively in the machine learning literature for approximately solving high dimensional and large sample statistical inference problems, particularly with Bayesian models. 
 In the next section, we briefly review approximate inference in a general setting.

\section{APPROXIMATE INFERENCE}\label{sec:VAE}
 Consider an ensemble of $n$ observations $\nY:=\{Y_1,Y_2,\ldots,Y_n\}$, where each $Y_i \in \mathcal A$, an  arbitrary topological space, and represents the available dataset. Each $Y_i$ induces a distribution $P_i$ that lies in some space of measures $\mathcal{P}$. 
The inference problem is to estimate a distribution over the sequence of unknown data generating distributions $\{P_1,P_2,\ldots,P_n\}$ given the observations $\nY$. In the Bayesian inference setting, we assume the existence of a sequence of `prior' distributions $\{\Pi_1,\Pi_2,\ldots,\Pi_n\} \in \mathcal{P}\times \cdots \times \mathcal P =: \bigotimes_{n}\mathcal{P}$. Subsequently, $Y_i\in\nY$ is assumed to follow a {\it generative model} defined in the following hierarchical manner:
    (i) Generate $P_i \sim \Pi_i$ $i\in \{1,2,\ldots n\}$; then
    (ii) sample $Y_i\sim P_i$.
 Now, using Bayes rule,  observe that for any subset $B\subseteq \bigotimes_{n}\mathcal{P}$ the `posterior' distribution satisfies
\begin{align}
    \Pi_n(B|\nY) = \frac{\int_{B}\prod_{i=1}^{n}d\Pi_i(P_i)P_i(Y_i)}{\int \prod_{i=1}^{n}d\Pi_i(P_i)P_i(Y_i)}.
    \label{eq:Post}
\end{align} 
Observe that we have assumed~$\nY$ forms an independent ensemble; we will continue with this assumption in the remainder of the paper. In most of the high (or possibly infinite) dimensional  settings computing this posterior distribution is intractable and consequently the problem of computing and performing statistical inference with the posterior is challenging. To address the intractability of the posterior, various sampling and optimization based methods have  been  proposed. We now describe a variational approach to do approximate inference that belongs to the latter category. In this framework, we first fix a class of measures $\mathcal{Q}_n \subseteq \bigotimes_n\mathcal{P}$ and then compute an approximation to the posterior~\eqref{eq:Post} in the family $\mathcal{Q}_n$ by optimizing a lower bound to the `model evidence' $\mathbb{P}(\nY):=\int \prod_{i=1}^{n}d\Pi_i(P_i)P_i(Y_i)$. Observe that for any sequence of measures $\{Q_i\}_{1\leq i \leq n} \in \mathcal{Q}_n$, Jensen's inequality implies  that
\begin{align}
\nonumber
   \log \mathbb{P}(\nY)
   \nonumber &\geq \int \prod_{i=1}^{n}dQ_i(P_i) \log \prod_{i=1}^{n} P_i(Y_i) - \int \prod_{i=1}^{n}dQ_i(P_i) \log \frac{\prod_{i=1}^{n}dQ_i(P_i)}{\prod_{i=1}^{n}d\Pi_i(P_i)}
   \\
   &= \sum_{i=1}^{n} \left[ \mathbb{E}_{Q_i}\left[\log P_i(Y_i) \right] - \frac{d Q_i}{d\Pi_i} (P_i) \right];
   \label{eq:ELBO}
\end{align}
in the machine learning literature the right hand side (RHS) in~\eqref{eq:ELBO} is popularly known as \textit{evidence lower bound} (ELBO). Observe that~\eqref{eq:vae} precisely corresponds to a single random variable in the sum on the RHS of~\eqref{eq:ELBO}, where the measure $Q_i$ corresponds to $P_{\Theta,k}$ and $\Pi_i$ corresponds to the measure $P_\Gamma$. In the variational framework, for a given sequence of prior distributions $\{\Pi_1,\Pi_2,\ldots,\Pi_n\}$ the ELBO is maximized over the  distribution in $\mathcal{Q}_n$ using stochastic gradient descent methods to find the best $\{Q_i\}_{\itn}$ in $\mathcal{Q}_n$. In particular, observe that the ELBO can be rewritten as
\[\textsc{ELBO} = -\textsc{KL} \left(\prod_{i=1}^{n}dQ_i(P_i) \Big\|d\Pi_n(P_1,P_2 \ldots P_n|\nY) \right)  + \log \bbP(\nY),\] 
where KL represents the Kullback-Leibler divergence. Therefore, the optimizer of ELBO is an approximation to the posterior  distribution as defined in~\eqref{eq:Post}. 

Deep latent models (DLMs)~\cite[Ch.\,19,\,20]{goodfellow} specialize this general presentation to the setting where the probability measures are parameterized by deep neural networks (DNNs). Variational autoencoders (VAEs)~\cite{Kingma2019} are an example of DLMs in the multivariate setting where the sequence of prior distributions are known only up to the parameters of an appropriately chosen DNN modeling these parameters. In the VAE literature this sequence of prior distributions are also known as \textit{decoders}. The approximating measures $\mathcal{Q}_n$, entitled \textit{encoders} in the VAE literature, are also parameterized using DNNs. Given the ensemble $\mathbf Y_n$, the DNN parameters of both the encoder and decoder are estimated using stochastic gradient descent (SGD). Our current setting, of course, is far more complicated than the VAE setting since the DNNs model the drift and diffusion coefficients of SDEs leading to a complicated training procedure, as we will see.

\section{DLMs FOR DSPPs}\label{sec:ps}
We assume access to $n$ independent and identically distributed (\iid) observations of a stochastic process $\{X(t), t\leq T\}$. In many service systems, such as hospitals and call centers, traffic counts are collected at fixed, regular intervals; for instance, in many large call centers, this is typically at intervals of length 30 seconds to 1 minute. As noted before, it has been observed~\cite{zhang2014scaling} that a DSPP with CIR-type ergodic diffusion process driving the intensity is an appropriate model of the traffic counts at operational time-scales (typically of the order of 10 minutes). The time interval $[0,T]$ in our model represents this operational time-scale.

For clarity of exposition, we will describe our method under two specific conditions: (i) the traffic counts are observed at the time epochs $T/2$ and $T$; and (ii) a single sample $n=1$. These can be extended to more observation instants and samples at the expense of a more burdensome notation, but our method will not change. We model the unknown stochastic intensity process by the SDE
\begin{align}~\label{eq:prior}
	dZ(t)=b(Z(t),t;\theta)dt+ \eta \sqrt{Z(t)}dW(t), \quad t\leq T
\end{align}
where $\{W(t),t\geq 0\}$ is the standard Brownian motion, $b(\cdot,t;\theta): C_b[0,T]\times [0,T]\mapsto \bbR$  is the drift and $\eta \sqrt{(\cdot)}$ with $\eta > 0$ is the diffusion coefficient. $C_b[0,T]$ denotes the space of all continuous and bounded function on the interval $[0,T]$. Here, the unknown drift function is modeled using a DNN parameterized by $\theta$, and to avoid getting bogged down in technical detail, we assume the existence of a strong solution to~\eqref{eq:prior}. For technical reasons we will, for now, assume that the diffusion coefficient is known. We denote the measure induced by the solution of this SDE as $P_{\theta}(\cdot)$; this corresponds to the `prior' measure in the previous section. The independent increments property implies that the joint distribution of the arrival count random vector $Y_1 := (X\left(\frac{T}{2}\right), X(T))$, conditional on the intensity process $Z_{0:T}$, can be expressed as:
   \begin{equation}\label{eqn:likelihood1}
   \begin{split}
   \bbP(Y_1 = (k_1,k_2) | Z_{0:T}) 
   &=\frac{e^{-\int_{0}^{\frac{T}{2}} Z(t) dt}\left(\int_{0}^{\frac{T}{2}} Z(t) dt\right)^{k_1}}{k_1!}  \frac{e^{-\int_{\frac{T}{2}}^{T} Z(t) dt}\left(\int_{\frac{T}{2}}^{T} Z(t) dt\right)^{k_2-k_1}}{(k_2-k_1)!}.
   \end{split} 
   \end{equation}

\subsection{A DLM for the Stochastic Intensity Process}
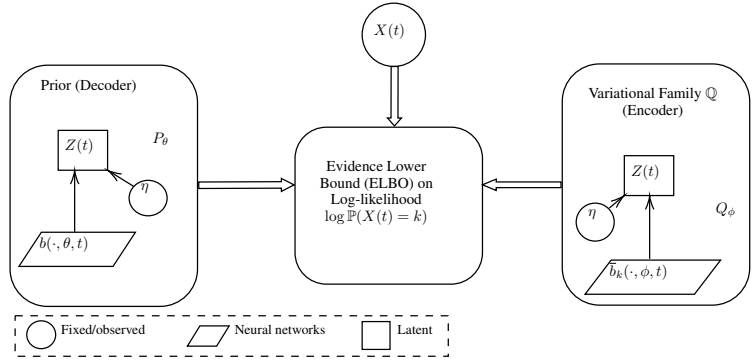
\begin{figure}
    \centering
    \tikzset{every picture/.style={line width=0.75pt}} 
    \resizebox{0.6\textwidth}{!}{%
\begin{tikzpicture}[x=0.75pt,y=0.75pt,yscale=-1,xscale=1]

\draw   (303,116) .. controls (303,102.19) and (314.19,91) .. (328,91) .. controls (341.81,91) and (353,102.19) .. (353,116) .. controls (353,129.81) and (341.81,141) .. (328,141) .. controls (314.19,141) and (303,129.81) .. (303,116) -- cycle ;
\draw   (63,191.5) -- (102,191.5) -- (102,223) -- (63,223) -- cycle ;
\draw    (76,272) -- (76,225) ;
\draw [shift={(76,223)}, rotate = 450] [color={rgb, 255:red, 0; green, 0; blue, 0 }  ][line width=0.75]    (10.93,-3.29) .. controls (6.95,-1.4) and (3.31,-0.3) .. (0,0) .. controls (3.31,0.3) and (6.95,1.4) .. (10.93,3.29)   ;
\draw   (472,264) .. controls (472,255.72) and (478.72,249) .. (487,249) .. controls (495.28,249) and (502,255.72) .. (502,264) .. controls (502,272.28) and (495.28,279) .. (487,279) .. controls (478.72,279) and (472,272.28) .. (472,264) -- cycle ;
\draw    (498,253) -- (508.46,244.28) ;
\draw [shift={(510,243)}, rotate = 500.19] [color={rgb, 255:red, 0; green, 0; blue, 0 }  ][line width=0.75]    (10.93,-3.29) .. controls (6.95,-1.4) and (3.31,-0.3) .. (0,0) .. controls (3.31,0.3) and (6.95,1.4) .. (10.93,3.29)   ;
\draw   (331.38,141.01) -- (331.27,181.75) -- (334.02,181.76) -- (328.51,187.99) -- (323.03,181.73) -- (325.78,181.74) -- (325.89,140.99) -- cycle ;
\draw   (250,213.8) .. controls (250,200.1) and (261.1,189) .. (274.8,189) -- (373.2,189) .. controls (386.9,189) and (398,200.1) .. (398,213.8) -- (398,288.2) .. controls (398,301.9) and (386.9,313) .. (373.2,313) -- (274.8,313) .. controls (261.1,313) and (250,301.9) .. (250,288.2) -- cycle ;
\draw    (530,293) -- (530,245) ;
\draw [shift={(530,243)}, rotate = 450] [color={rgb, 255:red, 0; green, 0; blue, 0 }  ][line width=0.75]    (10.93,-3.29) .. controls (6.95,-1.4) and (3.31,-0.3) .. (0,0) .. controls (3.31,0.3) and (6.95,1.4) .. (10.93,3.29)   ;
\draw   (461,168.6) .. controls (461,152.25) and (474.25,139) .. (490.6,139) -- (579.4,139) .. controls (595.75,139) and (609,152.25) .. (609,168.6) -- (609,300) .. controls (609,316.35) and (595.75,329.6) .. (579.4,329.6) -- (490.6,329.6) .. controls (474.25,329.6) and (461,316.35) .. (461,300) -- cycle ;
\draw   (510,211.5) -- (549,211.5) -- (549,243) -- (510,243) -- cycle ;
\draw   (119,245) .. controls (119,237.27) and (125.72,231) .. (134,231) .. controls (142.28,231) and (149,237.27) .. (149,245) .. controls (149,252.73) and (142.28,259) .. (134,259) .. controls (125.72,259) and (119,252.73) .. (119,245) -- cycle ;
\draw    (123,236) -- (103.7,224.05) ;
\draw [shift={(102,223)}, rotate = 391.76] [color={rgb, 255:red, 0; green, 0; blue, 0 }  ][line width=0.75]    (10.93,-3.29) .. controls (6.95,-1.4) and (3.31,-0.3) .. (0,0) .. controls (3.31,0.3) and (6.95,1.4) .. (10.93,3.29)   ;
\draw   (173.97,232.3) -- (239.22,232.04) -- (239.21,229.47) -- (249.22,234.56) -- (239.25,239.74) -- (239.24,237.17) -- (173.99,237.43) -- cycle ;
\draw   (398,234.5) -- (408.85,230) -- (408.85,232.25) -- (460,232.25) -- (460,236.75) -- (408.85,236.75) -- (408.85,239) -- cycle ;
\draw  [dash pattern={on 4.5pt off 4.5pt}] (29,338) -- (371,338) -- (371,370) -- (29,370) -- cycle ;
\draw   (38.18,353) .. controls (38.18,346.92) and (43.29,342) .. (49.59,342) .. controls (55.89,342) and (61,346.92) .. (61,353) .. controls (61,359.08) and (55.89,364) .. (49.59,364) .. controls (43.29,364) and (38.18,359.08) .. (38.18,353) -- cycle ;
\draw   (49.72,272) -- (124,272) -- (106.28,299) -- (32,299) -- cycle ;
\draw   (499.61,294) -- (586,294) -- (565.39,321) -- (479,321) -- cycle ;
\draw   (174.9,346) -- (198,346) -- (188.1,362) -- (165,362) -- cycle ;
\draw   (25,169.6) .. controls (25,153.25) and (38.25,140) .. (54.6,140) -- (143.4,140) .. controls (159.75,140) and (173,153.25) .. (173,169.6) -- (173,301) .. controls (173,317.35) and (159.75,330.6) .. (143.4,330.6) -- (54.6,330.6) .. controls (38.25,330.6) and (25,317.35) .. (25,301) -- cycle ;
\draw   (303,342.59) -- (325,342.59) -- (325,364.59) -- (303,364.59) -- cycle ;

\draw (311,106) node [anchor=north west][inner sep=0.75pt]  [xscale=0.9,yscale=0.9]  {$X( t)$};
\draw (67,197) node [anchor=north west][inner sep=0.75pt]  [xscale=0.9,yscale=0.9]  {$Z( t)$};
\draw (47,274) node [anchor=north west][inner sep=0.75pt]  [xscale=0.9,yscale=0.9]  {$b( \cdot ,\theta ,t)$};
\draw (480,253) node [anchor=north west][inner sep=0.75pt]  [xscale=0.9,yscale=0.9]  {$\eta $};
\draw (532.5,169.5) node  [xscale=0.9,yscale=0.9] [align=left] {\begin{minipage}[lt]{94.52000000000001pt}\setlength\topsep{0pt}
\begin{center}
Variational Family $\displaystyle \mathbb{Q}$ (Encoder)
\end{center}

\end{minipage}};
\draw (497,294.5) node [anchor=north west][inner sep=0.75pt]  [xscale=0.9,yscale=0.9]  {$\overline{b}_{k}( \cdot ,\phi ,t)$};
\draw (136,192) node [anchor=north west][inner sep=0.75pt]  [xscale=0.9,yscale=0.9]  {$P_{\theta }$};
\draw (581,247) node [anchor=north west][inner sep=0.75pt]  [xscale=0.9,yscale=0.9]  {$Q_{\phi }$};
\draw (514,217) node [anchor=north west][inner sep=0.75pt]  [xscale=0.9,yscale=0.9]  {$Z( t)$};
\draw (266,190) node [anchor=north west][inner sep=0.75pt]  [xscale=0.9,yscale=0.9] [align=left] {\begin{minipage}[lt]{78.13625pt}\setlength\topsep{0pt}
\begin{center}
\vspace{2em}
Evidence Lower 
Bound (ELBO) on \\ Log-likelihood $\displaystyle \log\mathbb{P}( X( t)=k) $
\end{center}

\end{minipage}};
\draw (127,232) node [anchor=north west][inner sep=0.75pt]  [xscale=0.9,yscale=0.9]  {$\eta $};
\draw (48,150) node [anchor=north west][inner sep=0.75pt]  [xscale=0.9,yscale=0.9] [align=left] {Prior (Decoder)};
\draw (64.11,343.77) node [anchor=north west][inner sep=0.75pt]  [xscale=0.9,yscale=0.9] [align=left] {{\footnotesize Fixed/observed}};
\draw (201,344) node [anchor=north west][inner sep=0.75pt]  [xscale=0.9,yscale=0.9] [align=left] {{\footnotesize Neural networks}};
\draw (329.65,343.84) node [anchor=north west][inner sep=0.75pt]  [xscale=0.9,yscale=0.9] [align=left] {{\footnotesize Latent}};
\end{tikzpicture}
}
    \caption{An illustration of the deep latent modeling framework.}
    \label{fig:Summary}
    \vspace{-1em}
\end{figure}

By definition, the variational family $\mathcal Q$ must consist of measures that are absolutely continuous with respect to the `prior' measure $P_\theta$. In our current setting, $\mathcal Q$ is the class of equivalent measures induced by the solutions of SDEs that have the same diffusion coefficient as~\eqref{eq:prior}. To be precise, consider the SDE
\begin{align}
d Z(t)&=\bar b_k( Z(t),t:\phi)dt+ \eta\sqrt{ Z(t)} dW(t), \text{ for $t\leq T$},
\label{eq:eq1}
\end{align}
where for each $k \in \{0,1,\ldots\}$ the drift function $\bar b_k(\cdot,\cdot;\phi)$ is modeled using a DNN with parameter $\phi$. We denote the measure induced by the solution of this SDE as $Q_{\phi}$.  Figure~\ref{fig:Summary} illustrates the use of deep latent models in defining measures $P_{\theta}$ and $Q_{\phi}$ and consequently ELBO.
Next, we derive the ELBO for the observation random vector $Y_1$. The proof,  omitted for space reasons, follows from Girsanov's theorem. 
\begin{theorem}\label{thm:ELBO}
Define $u_k(Z(t),t;\theta,\phi) :=(\eta \sqrt{Z(t)})^{-1} \bar{b}_k(Z(t),t;\phi)-b(Z(t),t;\theta)$ and suppose that $u_k$ satisfies a {\it strong} Novikov's condition,
\(
	\bbE\left[ \exp\left(\frac{1}{2}\int_0^T |u_k(Z(t), t; \theta, \phi)|^2 dt \right) \right] < +\infty ~\forall \theta,\phi.
\)
~Then, 
 \begin{equation}\label{eqn:hatW}
 \hat{W}_t:=\int_0^t u_k(Z(s),s;\theta,\phi)ds+W(t)
 \end{equation}
  is a Brownian motion w.r.t. $Q_{\phi}$ , $dZ(t)={b}(Z(t),t;\theta)dt+\eta \sqrt Z(t) d\hat{W}_t$, and
\begin{align}
    \log\bbP(Y_1 = (k_1,k_2)) &\geq \bbE_{Q_\phi}\left[      \log\bbP(Y_1 = (k_1,k_2)|Z_{0:T} )  - \frac{1}{2}\int_0^Tu_k^2(Z(s),s;\theta,\phi)ds  \right]:=\text{ELBO}.
    \label{eqn:ELBO}
\end{align}
\end{theorem}
Notice that we must assume that Novikov's condition holds for all possible parameterizations of the functions $\bar b$ and $b$. This is a strong condition that is satisfied for the class of DNNs that we work with in this paper, since the output of the DNN is bounded by definition. However, more analysis is required on sufficient conditions for DNNs to satisfy Novikov's condition. 

\subsection{TRAINING THE DLM}

Our objective is to train the neural networks $b(Z(t),t;\theta)$ and $ u_k(Z(t),t;\theta,\phi)$ by maximizing the ELBO. 
We fix $u_k(Z(t),t;\theta,\phi)$ to be a deterministic neural network defined as $ \tilde u(k,t;\beta)$ with parameters $\beta$. Combined with~\eqref{eqn:hatW}, this additional restriction imposed on $u_k(Z(t),t;\theta,\phi)$ ensures that the process $\hat W_t$ has independent increments. In the variational inference literature~\cite{blei2017variational}, this assumption is also known as the mean-field approximation, that is each partition of the unknown latent variable is independent of the other. A similar assumption on the latent  process was used in~\cite{tzen2019:neuralSDE}, where the authors call it a path-space analog of the mean-field approximation.  
Now substituting $u_k(Z(t),t;\theta,\phi)=\tilde u(k,t;\beta)$ in~\eqref{eqn:hatW} and from the observation $dZ(t)={b}(Z(t),t;\theta)dt+\eta \sqrt Z(t) d\hat{W}_t$, it follows that we can simulate the SDE using $  W(t)$ instead  of $ \hat W(t)$; that is,
\begin{equation}
dZ(t)=b(Z(t),t;\theta)dt+ \sqrt{Z(t)} \tilde u(k,t;\beta)dt+\sqrt {Z(t)} dW(t) \text{ and } Z(0)=0.
\label{eq:Var}
\end{equation}
We denote the measure induced by the above SDE as $Q_{\beta,\theta}$. For simplicity we fix $\eta=1$.

We use stochastic gradient descent (SGD) to maximize the objective in~\eqref{eqn:ELBO} to learn the unknown neural network parameters $\theta$ and $\beta$. In order to use SGD, we first need to generate sample paths of the latent process $Z(t)$~\eqref{eq:Var}, which we do using the  Euler-Maruyama discretization method. We partition the time interval $[0,T]$ in $N$ equal sub-intervals, denoted as $\{t_0,t_1,...t_N\}$, with $t_0=0$ and  $t_N=T$, set $Z(t_0)=Z(0)$, and simulate $\{Z(t_m)\}_{0\leq m \leq N}$ using the recursive equation
\begin{align*}
Z(t_{m+1})=Z(t_m)+b( Z(t_m),t_m,\theta)(t_{m+1}-t_m)&+\sqrt{Z(t_m)}\tilde u(k,t_m;\beta)(t_{m+1}-t_m) +\sqrt{Z(t_m)} \Delta W_m
\end{align*}
where $\{ \Delta W_m := W(t_{m+1})-W(t_{m})\}_{_{0\leq m < N}}$ are $N$ \iid standard Gaussians.  

In order to use SGD we also need to compute the gradient of the objective function~\eqref{eq:ELBO} with respect to the parameters $\theta$ and $\beta$.
Notice that the expectation in ELBO is with respect to the measure induced by SDE in~\eqref{eq:Var} denoted as $Q_{\beta,\theta}$. Observe that the only source of randomness in generating $Z(t)$ is from the Brownian motion $W(t)$, which  does not depend on either $\beta$ or $\theta$. Therefore we interchange the differential operator with respect to the parameters and the expectation in~\eqref{eqn:ELBO}. To make the dependence of $Z(t)$ on $\beta$ and $\theta$ explicit, we write $Z(t)$ as $Z^{\beta,\theta}(t)$.
In particular for given values of parameters $\theta$ and $\beta_{-j}$~(all components of parameter $\beta$ except $\beta^j$) observe that $\pb \bbE\left[      \log\bbP(Y_1 = (k_1,k_2)|Z^{\beta,\theta
   }_{0:T} )  - \frac{1}{2}\int_0^T \tilde u^2(k,s;\beta)ds  \right] =$
\begin{align}
\nonumber
   \nonumber
   \bbE\bigg[ \pb \log \bigg( \frac{e^{-\int_{0}^{\frac{T}{2}} Z^{\beta,\theta
   }(t) dt} \bigg(\int_{0}^{\frac{T}{2}} Z^{\beta,\theta
   }(t) dt\bigg)^{k_1}}{k_1!} &  \frac{e^{-\int_{\frac{T}{2}}^{T} Z^{\beta,\theta
   }(t) dt}\bigg(\int_{\frac{T}{2}}^{T} Z^{\beta,\theta
   }(t) dt\bigg)^{k_2-k_1}}{(k_2-k_1)!}\bigg)\\ & -\pb \int_0^T \tilde u^2(k,s;\beta)ds \bigg],
\end{align}
where we use the likelihood expression from~\eqref{eqn:likelihood1}.
Also note that, to avoid any confusion, we have omitted subscript $Q_{\beta,\theta}$ from $\bbE_{}[\cdot]$ above. Now applying straightforward product differentiation rule and subsequently interchanging the integral and $\pb$, would result into an expression requiring us to compute the derivative of the process $Z^{\beta,\theta
}(t)$ with respect to $\beta^j$. To compute the derivative process, it follows from~\cite[Theorem 3.1]{kunita1984stochastic} 
that under certain regularity condition on the drift and diffusion coefficient of the process $Z^{\beta,\theta
   }(t)$~\eqref{eq:Var} the derivative process $\pb Z^{\beta,\theta
   }(t)$ is the solution of the following SDE 
  \begin{align*}
\frac{\partial Z^{\beta,\theta
   }(t)}{\partial \beta^j}=\int_0^t &\left( \frac{\partial b(Z^{\beta,\theta}(s),s;\theta)}{\partial Z^{\beta,\theta
   }(s)}\frac{\partial Z^{\beta,\theta
   }(s)}{\partial\beta^j} + \frac{\tilde u(k,s;\beta)}{2\sqrt{Z^{\beta,\theta
   }(s)}}\frac{\partial Z^{\beta,\theta
   }(s)}{\partial \theta^j} 
   + \sqrt{Z^{\beta,\theta}(s)}\frac{\partial u_s}{\partial\beta^j} \right)ds 
   \\& +\int_0^t \left(\frac{1}{2\sqrt{Z^{\beta,\theta
   }(s)}}\frac{\partial Z^{\beta,\theta
   }(s)}{\partial\beta^j}\right)dW_s \text{ and  } \frac{\partial Z^{\beta,\theta
   }(0)}{\partial \beta^j} =0.
\end{align*}
We simulate the derivative process above using the Euler-Maruyama method in a similar manner as we did for $Z^{\beta,\theta}(t)$. Lastly, we use a similar procedure  to generate the derivative of the latent process $Z^{\beta,\theta}(t)$ with respect to a component of $\theta$ for given values of the other parameters. We omit this for space reasons.
\section{NUMERICAL EXPERIMENTS}
We conducted a number of simple experiments to demonstrate both the in- and out-of-sample performance of the DLM. We start by describing the setting for the experiments. The code is written in Matlab with the Deep Learning Toolbox. The computation and space complexity of this method can be found in \cite{tzen2019:neuralSDE}. In our specific case, the time complexity for each iteration of the gradient update is $\mathcal{O}\left(N((k+n)T(b)+T(\tilde u))\right)$, where $k$ and $n$ are number of parameters in $\beta$ and $\theta$, respectively, $T(f)$ is the time complexity for computing $f$, and $N$ is the number of time steps in the time discretization.

\subsection{Setting}
Observe that training the neural network by maximizing the ELBO entails solving a stochastic optimization problem in~\eqref{eq:ELBO}. We use a sample average approximation (SAA) of~\eqref{eq:ELBO} for which we simulate $m$ independent sample paths of $(Z(t) : t\in [0,T])$: 
\begin{equation}\label{eqn:empirical_ELBO}
\frac{1}{mn} \sum_{i=1}^n\sum_{j=1}^m \left[   \log\left( P(X^i(T/2)=k^i_1,X^i(T)=k^i_2|Z^j_{0:T}  \right) - \frac{1}{2}\int_0^T\tilde u^2(k,s;\beta)ds  \right].
\end{equation}
We integrate the SDEs using Euler-Maruyama discretization noted in the previous section. 
The architecture of the neural networks is
\begin{itemize}
    \item $b(Z(t),t;\theta):R^2\rightarrow R$ is a feedforward neural network with 20 fully connected layers of size 10. The activation function is chosen as $tanh$. The inputs are time epoch and the current intensity.
    \item $\tilde u(k,t;\beta):R^2\rightarrow R$ is also a feedforward neural network with 20 fully connected layers of size 10. The activation function is chosen as $tanh$. The inputs are time epoch and the state at time $T$.
\end{itemize}
Notice that unlike \cite{li;ScalableNeuralSDE}, we do not require any specific architecture on the Neural network. One can always tune the hyperparameters to find as good or even better architecture.

We assume that the true latent intensity process is a standard CIR process:
\begin{equation}\label{eqn:theoretical_model}
dZ(t)=0.3 (80-Z(t))dt+\eta\sqrt{Z(t)}dW(t),
\end{equation}
where $Z(0)=5$ and $\eta\in [0,1]$ is the `noise magnitude' of the model. Observe that if $\eta = 0$, the intensity process is the solution of an ordinary differential equation, and the arrival process is a NHPP. We set the simulation horizon to be $T=4$, and uniformly partition the interval $[0,T]$ into the grid $\mathcal{P}=\lbrace t_1,t_2,..,t_M \rbrace$ with $t_{k+1}-t_k=1/15$, $t_1=0$ and $t_M=4$. The training data consists of $n=200$ sample paths of the DSPP  generated using the theoretical model \eqref{eqn:theoretical_model}. This data is further divided into `mini-batches' of size 10 and then fed into the Adam solver~\cite{kingma2014adam}. We run the code for 35 epochs (350 gradient updates in total). The learning rate for $b(Z(t),t;\theta)$ and $\tilde u(k,t;\beta):R^2\rightarrow R$ are both set to be 0.01. \par

We compare our method against the piece-wise linear maximum likelihood estimate (MLE) of the intensity assuming the traffic model is an NHPP, developed in~\cite{Zeyu:pl}. This estimator is quite robust when the objective is to predict a mean performance metric. While it can be very inaccurate in predicting higher moments, betraying the fact that the MLE is computed assuming no correlation structure in the count process, we believe the relative simplicity and the fact that the computation of mean performance metrics are frequently the focus of performance analysis make it a useful reference. 
As noted before, our experiment will use arrival counts at time $T/2$ and $T$, and therefore it suffices to consider a two piece linear estimator in this experiment by maximizing the likelihood function,
\begin{equation*}
\mathcal{L}_n (\hat Z(t))=\frac{1}{n} \sum_{i=1}^n \left( X^i(T/2)\int_0^\frac{T}{2} \hat Z(t) dt+(X^i(T)-X^i(T/2))\int_\frac{T}{2}^T \hat Z(t) dt \right)-\int_0^T \hat Z(t) dt,
\end{equation*}
which follows from display (2) in~\cite{Zeyu:pl}.

\subsection{Estimating the Intensity Process}\label{subsec:DSPP_arrival}
\begin{figure}[htb]
     \centering
     \begin{subfigure}[b]{0.49\textwidth}
         \centering
         \includegraphics[trim={0 0 0 0.8cm},clip,height=0.5\textwidth,width=0.95\textwidth]{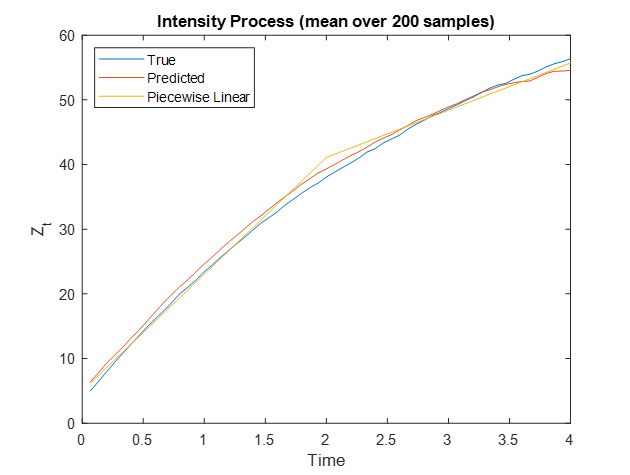}
         \caption{Intensity Process}
         \label{fig:Intensity Process}
     \end{subfigure}
     \hfill
     \begin{subfigure}[b]{0.49\textwidth}
         \centering
         \includegraphics[trim={0 0 0 0.8cm},clip,height=0.5\textwidth,width=0.95\textwidth]{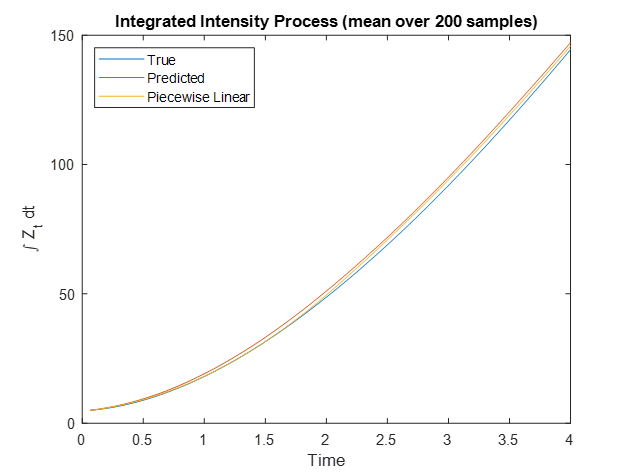}
         \caption{Integrated Intensity Process}
         \label{fig:Integrated Intensity Process}
     \end{subfigure}
        \caption{Learning result for Model \eqref{eqn:theoretical_model}.\label{fig:DSPP}}
        \vspace{-2em}
\end{figure}

Our first experiment focuses on the estimation of the `true' latent intensity model~\eqref{eqn:theoretical_model} when $\eta=1$. Figure \ref{fig:DSPP}(a) shows the results of an in-sample estimation of the average intensity (computed using 200 training samples of the `true' model). Observe that both our method (`predicted') and the piece-wise linear model estimate the mean intensity process quite accurately. Figure~\ref{fig:DSPP}(b) shows that the estimated mean integrated intensity, too, is almost identical to the `true' model in either model. This is unsurprising: recall that the Poisson count distribution in the ELBO~\eqref{eqn:ELBO} is a function of the integrated intensity, and this plays a crucial role in constraining the estimation problem. 


\subsection{Performance Prediction in an Infinite Server Queue.}\label{subsc:InfiniteServerQueue}
In the second experiment, we focus on an out-of-sample performance prediction task for an infinite server queue. Specifically, we conduct `run-through' experiments where traffic generated from a DSPP using estimated intensity processes is used as an input to a simulation of an infinite server queue.  We start with a $G_t/M/\infty$ queue, where traffic is generated using the theoretical model \eqref{eqn:theoretical_model}, and the deep latent and piece-wise linear models estimated in the previous section. 

We generate 500 sample paths of the number of occupied servers over $[0,T]$ with service rate $\mu=2$. Observe from Table \ref{table:GMqueue} that the estimated DLM gives a reasonable inference of both the mean and variance of the number of occupied servers at $\frac{T}{2}$ and $T$. Note that the variance is roughly in the ballpark of the variance of the `true' model as estimated from the test dataset. On the other hand, the piece-wise linear model, underestimates the variance quite significantly.\par
\begin{table}[htb]
\caption{Simulation of the number of occupied servers for a $G_t/M/\infty$ queue.\label{table:GMqueue}}
\resizebox{\textwidth}{!}{
\begin{tabular}{|l|l|l|l|l|l|l|}
\hline
\multicolumn{1}{|c|}{\multirow{2}{*}{Number   of Occupied Servers}} & \multicolumn{2}{c|}{Test}       & \multicolumn{2}{c|}{DLM}       & \multicolumn{2}{c|}{PL}         \\ \cline{2-7} 
\multicolumn{1}{|c|}{}                                              & T/2            & T              & T/2            & T              & T/2            & T              \\ \hline
Mean$\pm$CI                                                               & 30.73$\pm0.68$ & 72.97$\pm$1.33 & 31.18$\pm$0.75 & 71.75$\pm$1.51 & 31.88$\pm$0.60 & 73.14$\pm$1.05 \\ \hline
Variance                                                            & 61.2           & 229.93         & 73.25          & 296.2          & 47.54          & 143.76         \\ 
CI                                                   & 54.26   69.56  & 203.87  261.36 & 64.94   83.26  & 262.62  336.68 & 42.16   54.04  & 127.47  163.41 \\ \hline
\end{tabular}}
\end{table}
Next, we repeat the previous experiment on a $G_t/G/\infty$ system, with Erlang distributed service times, parameterized by $\lambda=6$ and $k=3$ (implying the mean  service time remains $\frac{1}{2}$). The simulation is summarized in Table \ref{table:GEkqueue}. Again, the DLM model makes acceptable predictions. We note that while the DLM model tends to predict higher variance, and the estimates tend to have a larger confidence interval, we conjecture that the accuracy of the predictions can be improved with a more appropriate choice of neural network size and more Monte Carlo samples in the SAA approximation to the ELBO (recall we have used $m=5$ throughout).

\begin{table}[htb]
\caption{Simulation of the number of occupied servers for a $G_t/G/\infty$ queue.\label{table:GEkqueue}}
\resizebox{\textwidth}{!}{
\begin{tabular}{|l|l|l|l|l|l|l|}
\hline
\multicolumn{1}{|c|}{\multirow{2}{*}{Number   of Occupied Servers}} & \multicolumn{2}{c|}{Test}       & \multicolumn{2}{c|}{DLM}       & \multicolumn{2}{c|}{PL}         \\ \cline{2-7} 
\multicolumn{1}{|c|}{}                                              & T/2            & T              & T/2            & T              & T/2            & T              \\ \hline
Mean$\pm$CI                                                                & 37.34$\pm$0.67 & 85.5$\pm$1.39  & 37.57$\pm$0.72 & 84.36$\pm$1.51 & 38.18$\pm$0.57 & 87.35$\pm$1.04 \\ \hline
Variance                                                            & 58.89          & 253.57         & 68.05          & 298.73         & 42.47          & 140.33         \\ 
CI                                                  & 52.21   66.94  & 224.83  288.23 & 60.33   77.35  & 338.73  264.22 & 37.65   48.27  & 124.42 159.51  \\ \hline
\end{tabular}}
\end{table}
\vspace{-1em}
\subsection{Impact of the Noise Factor}\label{subsec:smalleta}
\begin{figure}[htb]
     \centering
     \begin{subfigure}[b]{0.49\textwidth}
         \centering
         \includegraphics[trim={0 0 0 0.8cm},clip,height=0.6\textwidth,width=0.95\textwidth]{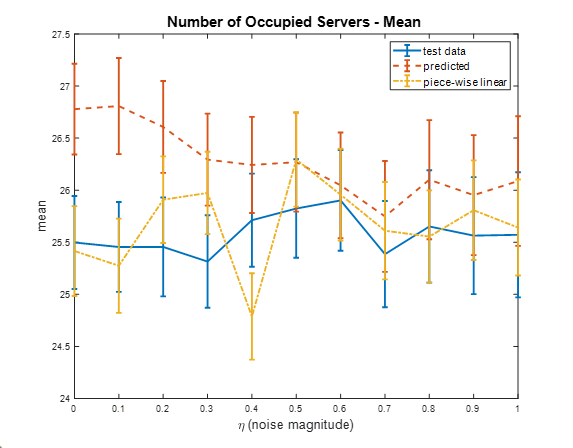}
         \caption{Mean}
         \label{fig:Mean}
     \end{subfigure}
     \hfill
     \begin{subfigure}[b]{0.49\textwidth}
         \centering
         \includegraphics[trim={0 0 0 0.8cm},clip,height=0.6\textwidth,width=0.95\textwidth]{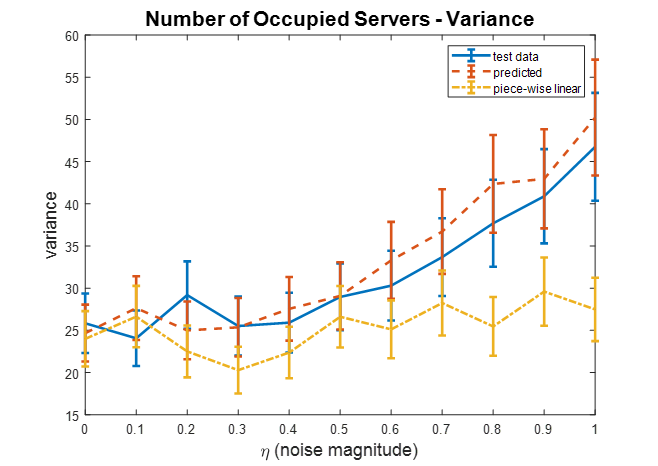}
         \caption{Variance}
         \label{fig:Variance}
     \end{subfigure}
        \caption{Statistical inference on the number of occupied servers as a function of the noise factor.}
        \vspace{-1em}
\end{figure}

The previous experiments demonstrate that the DLM model is robust on both mean and variance prediction tasks. To further explore this, in this experiment we demonstrate how the DLM predictions change when the `noise factor' $\eta$ in~\eqref{eqn:theoretical_model} increases from 0 to 1; here, $\eta = 0$ (formally) corresponds to a deterministic intensity and $\eta > 0$ to increasing levels of stochasticity in the intensity model. We conducted the same `run-through' experiment from the previous section on a $G_t/M/\infty$ queue, albeit with different estimated traffic models under the different $\eta$ factors. 
While this is might appear surprising, recall that the mean number of occupied servers under the `annealed' measure (i.e., averaged over the stochastic intensity) of an infinite server queue with DSPP traffic depends only the mean intensity function. The PL estimate, even though it is based on the `quenched' (i.e., conditioned on the intensity) measure, accurately estimates the mean intensity when averaged over the individual sample paths. For larger $\eta$, we observe that the DLM makes reasonable predictions on the mean number of occupied servers. On the other hand, Figure \ref{fig:Variance} shows that when $\eta$ increases the DLM model significantly outperforms the piece-wise linear model in predicting the variance of the number of occupied servers. This is due to the fact that the DLM estimates the annealed measure of the traffic model, while the PL model only estimates the quenched measure.

\subsection{Estimating a Nonhomogeneous Poisson Intensity}
We demonstrate the robustness of our method on estimating the intensity of a NHPP, with deterministic intensity. Consider an intensity function that is the solution of the ordinary differential equation (ODE) $\dot{Z}(t) = a(b-Z(t))$ with $a = 0.3$ and $b=80$. Let $d$ be the number of time intervals (or `pieces') in the regressors, representing the number of degrees of freedom. We compare our method, using intensity process $dZ(t) = a(b-Z(t)) dt + d^{-1/2}\sqrt{Z(t)}dW(t)$, with the piecewise linear estimator~\cite{Zeyu:pl} and the nonparametric `Gaussianization machine' method from~\cite{cai2019gaussianization} (`GRP' in the table below). GRP uses a variance stabilizing transformation of the Poisson counts and Gaussian process regression on the transformed variables. Table~\ref{table:discretization} shows that our method is significantly better than GRP even with 50 degrees of freedom.

\begin{table}[htb]
 \caption{Traffic count prediction for NHPP.\label{table:discretization}}
 \resizebox{\textwidth}{!}{
\begin{tabular}{|c|c|c|c|c|c|c|c|c|}
\hline
\multicolumn{1}{|l|}{} & \multicolumn{4}{c|}{Mean} & \multicolumn{4}{c|}{Variance} \\ \hline
\multicolumn{1}{|c|}{d} &
  \multicolumn{1}{c|}{Test} &
  \multicolumn{1}{c|}{Predicted} &
  \multicolumn{1}{c|}{Piecewise Linear} &
  \multicolumn{1}{c|}{GRP} &
  \multicolumn{1}{c|}{Test} &
  \multicolumn{1}{c|}{Predicted} &
  \multicolumn{1}{c|}{Piecewise Linear} &
  \multicolumn{1}{c|}{GRP} \\
  \hline
\multirow{2}{*}{2} &
  \multirow{2}{*}{146.46} &
  \multirow{2}{*}{146.73$\pm$1.09} &
  \multirow{2}{*}{149.30$\pm$1.10} &
  \multirow{2}{*}{116.51$\pm$5.79} &
  \multirow{2}{*}{146.46} &
  \multirow{2}{*}{\begin{tabular}[c]{@{}c@{}}156.17\\      {[} 138.46  177.51{]}\end{tabular}} &
  \multirow{2}{*}{\begin{tabular}[c]{@{}c@{}}158.93\\      {[}140.91  180.65{]}\end{tabular}} &
  \multirow{2}{*}{\begin{tabular}[c]{@{}c@{}}$4.34*10^3$\\      {[}$3.85*10^3$ $4.94*10^3$ {]}\end{tabular}} \\
                       &         &        &        &          &   &       &   &      \\ \hline
\multirow{2}{*}{10} &
  \multirow{2}{*}{146.46} &
  \multirow{2}{*}{145.93$\pm$1.11} &
  \multirow{2}{*}{148.32$\pm$1.10} &
  \multirow{2}{*}{128.45$\pm$4.07} &
  \multirow{2}{*}{146.46} &
  \multirow{2}{*}{\begin{tabular}[c]{@{}c@{}}160.91\\      {[}142.67  182.90{]}\end{tabular}} &
  \multirow{2}{*}{\begin{tabular}[c]{@{}c@{}}159.50\\      {[}141.42  181.30{]}\end{tabular}} &
  \multirow{2}{*}{\begin{tabular}[c]{@{}c@{}}$2.15*10^3$\\      {[}$1.91*10^3$ $2.44*10^3${]}\end{tabular}} \\
                       &         &        &        &          &   &       &   &      \\ \hline
\multirow{2}{*}{20} &
  \multirow{2}{*}{146.46} &
  \multirow{2}{*}{146.25$\pm$1.07} &
  \multirow{2}{*}{146.87$\pm$1.04} &
  \multirow{2}{*}{140.93$\pm$3.18} &
  \multirow{2}{*}{146.46} &
  \multirow{2}{*}{\begin{tabular}[c]{@{}c@{}}148.33\\      {[}131.52  168.60{]}\end{tabular}} &
  \multirow{2}{*}{\begin{tabular}[c]{@{}c@{}}142.44\\      {[}126.30  161.91{]}\end{tabular}} &
  \multirow{2}{*}{\begin{tabular}[c]{@{}c@{}}$1.31*10^3$\\      {[}$1.16*10^3 1.48*10^3${]}\end{tabular}} \\
                       &         &        &        &          &  &    &    &         \\ \hline
\multirow{2}{*}{50} &
  \multirow{2}{*}{146.46} &
  \multirow{2}{*}{146.87$\pm$1.06} &
  \multirow{2}{*}{150.16$\pm$1.11} &
  \multirow{2}{*}{146.03$\pm$2.26} &
  \multirow{2}{*}{146.46} &
  \multirow{2}{*}{\begin{tabular}[c]{@{}c@{}}146.06\\      {[}129.50  166.02{]}\end{tabular}} &
  \multirow{2}{*}{\begin{tabular}[c]{@{}c@{}}161.52\\      {[}143.21  183.59{]}\end{tabular}} &
  \multirow{2}{*}{\begin{tabular}[c]{@{}c@{}}662.51\\      {[}587.41  753.06{]}\end{tabular}} \\
                       &         &        &        &          &  &    &    &        \\ \hline
\end{tabular}}
\end{table}

\subsection{Estimating the Diffusion Coefficient}
We presented our method under the assumption that the diffusion coefficient is known. However,~\cite{zhang2014scaling} argue that the model in~\eqref{eq:cir} is appropriate for modeling the stochastic arrival intensity in a range of service systems. Estimating this model necessitates consideration of the situation where both the diffusion and drift function are unknown. In this section we present numerical results showing that our method can work even in this situation. We assume that $\theta=80$, $\eta=1$ and $\alpha=\frac{1}{4}$ in the theoretical/true model. Table~\ref{table:Diffusion_Drift} below summarizes the results of the experiment.

We model the diffusion function by another neural network $\sigma(Z(t),t;\hat\theta)$ with the same structure as $b(Z(t),t;\theta)$. Notice that the only difference in the training framework is that, following the definitions in Section \ref{sec:VAE}, we must use $\sigma(Z(t),t;\hat\theta)u_k(Z(t),t;\theta,\phi)=\bar{b}(Z(t),t;\phi)-b(Z(t),t;\theta)$ to define $u_k(Z(t),t;\theta,\phi)$. 

\begin{table}[htb]
\caption{Traffic count prediction for DSPP with learnt diffusion coefficient.\label{table:Diffusion_Drift}}
\resizebox{\textwidth}{!}{
\begin{tabular}{|l|l|l|l|l|l|l|}
\hline
\multicolumn{1}{|c|}{\multirow{2}{*}{Traffic Counts}} & \multicolumn{2}{c|}{Test}       & \multicolumn{2}{c|}{DLM}       & \multicolumn{2}{c|}{PL}        \\ \cline{2-7} 
\multicolumn{1}{|c|}{}                                  & T/2            & T              & T/2            & T              & T/2           & T              \\ \hline
Mean$\pm$CI                                                    & 94.93$\pm$2.35    & 246.9$\pm$5.52    & 93.1$\pm$2.17     & 240.13$\pm$6.99   & 97.07$\pm$0.83   & 248.42$\pm$1.35   \\ \hline
Variance                                                & 285.91         & 1569.52        & 243.26         & 2518.06        & 90.62         & 237.01         \\
CI                                      & 253.50  324.99 & 1391.62  1784.6 & 215.68  276.50 & 2232.61 2862.2 & 80.35  103.01 & 210.15  269.41 \\ \hline
\end{tabular}}
\end{table}
\vspace{-1em}
\section{Conclusions and Commentary}
 This paper presents a versatile computational method for estimating the stationary, ergodic stochastic intensity of a DSPP. We demonstrate our method by in-sample estimation of the intensity and out-of-sample run-through simulation experiments, both of which demonstrate accuracy of our method. We believe that the method presented in this paper demonstrates how machine learning can help enhance simulation and modeling, in the spirit of the observation made by Peter W. Glynn in his {\it Titans of Simulation} keynote lecture at the Winter Simulation Conference in 2019~\cite{PGTalk}. 
In future work we intend to extend our method to jump Markov intensities and self-exciting traffic models (such as the Hawkes process). 
In on-going work we are developing large sample statistical analyses of DLMs on general measure spaces (including asymptotic consistency and central limit theorems), and will be presented in future papers.



\footnotesize
\bibliographystyle{unsrt}

\bibliography{demobib}
\section*{AUTHOR BIOGRAPHIES}
\noindent {\bf Ruixin Wang } is currently is Ph.D. student in the School of Industrial Engineering, specializing in Operations Research. His research interests lie in approximate dynamic programming, electricity market modeling and stochastic simulation. His email address is \href{wang2252@purdue.edu}{wang2252@purdue.edu}.\\

\noindent {\bf PRATEEK JAISWAL} is a Ph.D. candidate in the School of Industrial Engineering at Purdue University. His research interests are in machine learning and stochastic optimization. His e-mail address is \href{jaiswalp@purdue.edu}{jaiswalp@purdue.edu}.\\

\noindent {\bf HARSHA HONNAPPA} is an assistant professor in the School of Industrial Engineering at Purdue University. His research interests are in applied probability, game theory, and machine learning. He is a member of INFORMS, IEEE, and SIAM, and serves as an associate editor for Operations Research and Operations Research Letters. His email address is \href{honnappa@purdue.edu}{honnappa@purdue.edu}.\\

\end{document}